\documentclass[sigconf]{acmart}

\AtBeginDocument{%
  }

\setcopyright{acmcopyright}
\copyrightyear{2018}
\acmYear{2018}
\acmDOI{XXXXXXX.XXXXXXX}

\acmConference[AIPR 2024]{7th International Conference on Artificial Intelligence and Pattern Recognition (AIPR)}{September 20--22,
  2024}{Xiamen, China}
\acmISBN{979-8-4007-1717-8/24/09}

\begin{document}

\title{Normality Addition via Normality Detection in Industrial Image Anomaly Detection Models}

\author{Jihun Yi}
\affiliation{%
\department{Department of Electrical and Computer Engineering}
  \institution{Seoul National University}
  \city{Seoul}
  \country{Korea}
}
\email{t080205@snu.ac.kr}

\author{Dahuin Jung}
\affiliation{%
\department{School of Computer Science and Engineering}
  \institution{Soongsil University}
  \city{Seoul}
  \country{Korea}}
\email{dahuin.jung@ssu.ac.kr}

\author{Sungroh Yoon}
\authornote{Corresponding author}
\affiliation{%
\department{Department of Electrical and Computer Engineering and Interdisciplinary Program in AI}
  \institution{Seoul National University}
  \city{Seoul}
  \country{Korea}}
\email{sryoon@snu.ac.kr}

\renewcommand{\shortauthors}{Yi et al.}

\newcommand{\xvec}{\mathbf{x}}
\newcommand{\Xvec}{\mathbf{X}}
\newcommand{\pvec}{\mathbf{p}}
\newcommand{\Atheta}{\mathcal{A}_{\theta}}
\newcommand{\mmap}{\mathcal{M}}
\newcommand{\Loss}{\mathcal{L}}

\newcommand{\textblue}{\textcolor{blue}}
\newcommand{\textred}{\textcolor{red}}
\newcommand{\xmark}{\text{\ding{55}}}
\newcommand{\cmark}{\text{\ding{51}}}

\newcommand{\etal}{\emph{et al.}}
\newcommand{\cond}{\,|\,}
\newcommand{\ie}{\textit{i}.\textit{e}.}
\newcommand{\eg}{\textit{e}.\textit{g}.}
\newcommand{\versus}{\textit{v}.\textit{s}.}
\begin{abstract}
  The task of image anomaly detection (IAD) aims to identify deviations from normality in image data.
These anomalies are patterns that deviate significantly from what the IAD model has learned from the data during training. 
However, in real-world scenarios, the criteria for what constitutes normality often change, necessitating the reclassification of previously anomalous instances as normal.
To address this challenge, we propose a new scenario termed ``normality addition,'' involving the post-training adjustment of decision boundaries to incorporate new normalities.
To address this challenge, we propose a method called Normality Addition via Normality Detection (NAND), leveraging a vision-language model.
NAND performs normality detection which detect patterns related to the intended normality within images based on textual descriptions.
We then modify the results of a pre-trained IAD model to implement this normality addition.
Using the benchmark dataset in IAD, MVTec AD, we establish an evaluation protocol for the normality addition task and empirically demonstrate the effectiveness of the NAND method.

\end{abstract}

\ccsdesc[500]{Computing methodologies~Visual inspection}
\ccsdesc[300]{Computing methodologies~Scene anomaly detection}
\ccsdesc[100]{Computing methodologies~Image representations}

\keywords{anomaly detection, vision inspection, vision language model}
\begin{teaserfigure}
\centering
  \includegraphics[width=0.9\textwidth]{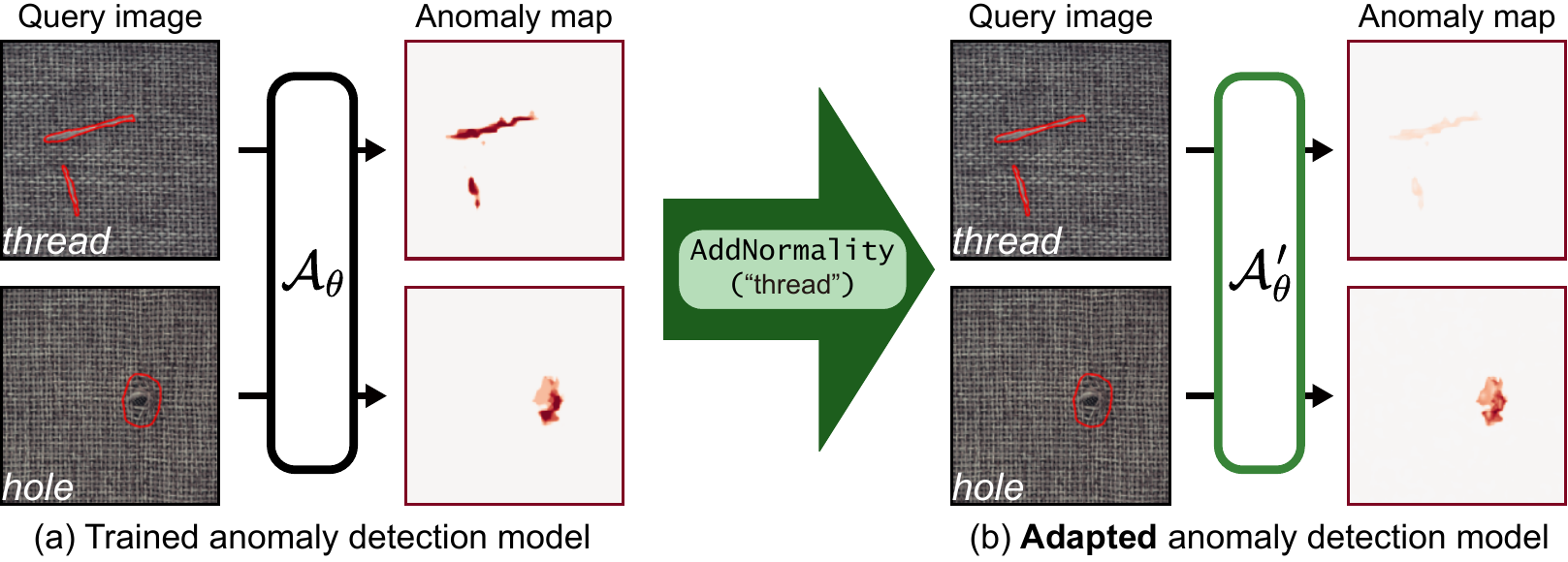}
  \caption{\textbf{Normality addition to a trained image anomaly detection model.} (a) The original anomaly detection model, denoted as $\mathcal{A}_\theta$, identifies both thread and hole in the carpet images as anomalies.
  The proposed \textit{normality addition} process adds the normality of ``thread'' to the model.
  (b) The resulting anomaly detection model classifies thread as normal, while leaving a hole as abnormal.
  }
  \label{fig:teaser} \label{fig:na_problem}
\end{teaserfigure}

\maketitle

\section{Introduction}
Advancements in deep learning have led to the development of numerous image anomaly detection models, notably for defect detection in manufacturing industries~\cite{roth2022towards,sabokrou2018adversarially,yan2021learning}.
These models leverage deep neural networks to identify anomalies in images, enhancing the quality control processes of manufacturers.
With increasing demand for process automation, research interest in anomaly detection continues to rise~\cite{ucad,inctrl}.

Previous research~\cite{sabokrou2018adversarially,yan2021learning} has predominantly focused on building anomaly detection models, with little attention given to modifying the learned knowledge of these models after training. 
However, in practical scenarios, there is often a need to adjust the decision boundary of anomaly detection models during operation.
For example, changes in manufacturing processes or environments may require recalibrating the model's sensitivity to anomalies.
Additionally, there may be instances where the criteria for determining anomaly status in quality control change, prompting the reclassification of previously labeled abnormalities as normal.

So far, addressing such label shifts required collecting new data and either retraining the model from scratch or fine-tuning it.
However, this approach is both time-consuming and resource-intensive.
Acquiring sufficient training data can be especially challenging in industries where anomalies are rare.
Moreover, starting from scratch with model retraining may not be viable when a quick adaptation is necessary.
Hence, we propose the scenario of ``normality addition'' in image anomaly detection problem, which demands swift adaptation of model to add new normality.
The example of normality addition is illustrated in Fig.~\ref{fig:teaser}.

Furthermore, we introduce a text-guided adaptation method based on vision-language model, named as Normality Addition via Normality Detection (NAND), to solve this problem.
NAND enables post-hoc modification of existing models without the need for extensive retraining.
NAND performs detection of the adding normality in query image, subsequently modifying the outputs of the existing anomaly detection models accordingly.
The contributions of this paper can be summarized as follows:
\begin{enumerate}
\item We propose the problem of normality addition in image anomaly detection and present an evaluation protocol of the task.
\item We introduce a text-guided normality addition method, named as NAND, which uses vision-language model to incorporate new normality provided as text form.
\item We empirically demonstrate the feasibility of normality addition by using the proposed NAND method.
\end{enumerate}

\section{Background}
\subsection{Vision-Language Models (VLMs)}
The emergence of vision-language models (VLMs), starting from CLIP~\cite{clip2021icml}, has revolutionized the integration of vision and text data.
These multi-modal models have the capability to process and understand both images and text simultaneously.
CLIP achieves this by training a multi-modal encoder on paired image-text data using contrastive learning, enabling the establishment of a shared embedding space for both modalities.
Zero-shot classification is made possible by finding the text prompt containing the class names closest to the query image in the common embedding space, and the description in Fig.~\ref{fig:clip_and_coop}(a) illustrates this process.
More formally, for a given image $x$ and a set of $C$ class names, $\left\{c_i\right\}_{i=1}^C$, zero-shot classification result using CLIP can be expressed as Eq.~\ref{eq:clip}:

\begin{figure*}[t]
    \centering
    \includegraphics[width=\linewidth]{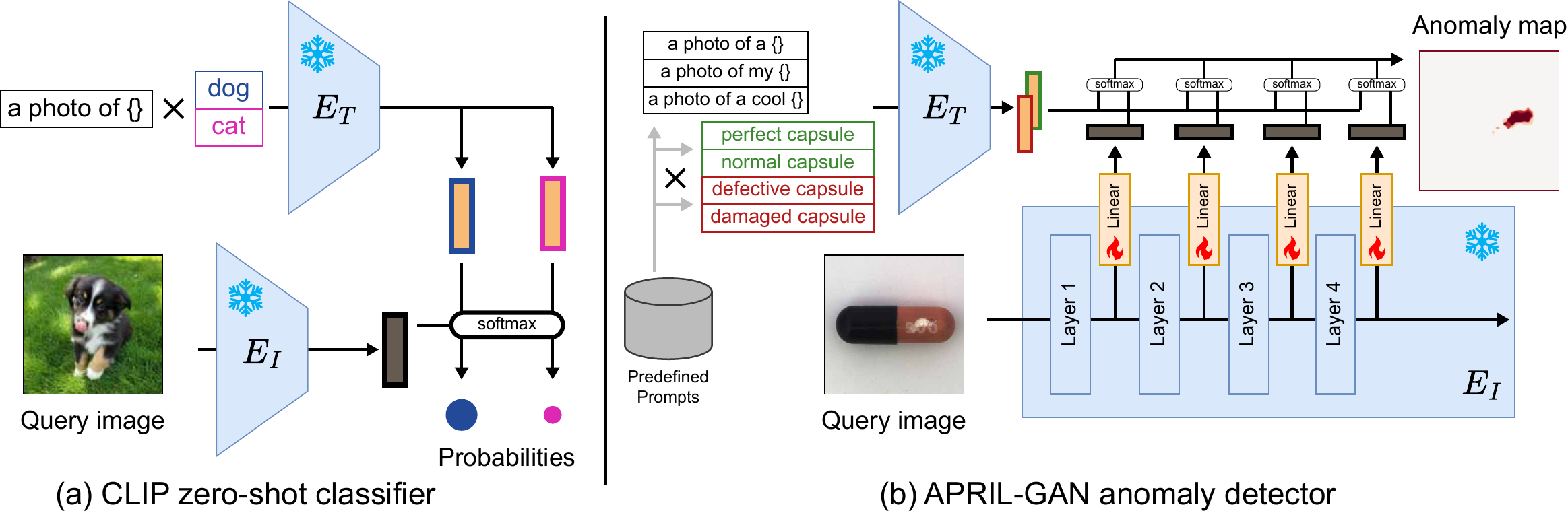}
    \vspace{-20pt}
    \caption{\textbf{A schematic description of (a) CLIP~\cite{clip2021icml} and (b) APRIL-GAN~\cite{aprilgan}.}}
    \vspace{-10pt}
    \label{fig:clip_and_coop}
\end{figure*}

\begin{equation} \label{eq:clip}
    P_{CLIP}(c_i|x) = \text{softmax} \left( E_I(x), \left\{  E_T \left(t(c_j) \right) \right\}_{j=1}^{C} \right)_i, 
\end{equation}

where $t(\cdot)$ denotes the prompt-generating function, and $E_T$ and $E_I$ represent the CLIP text and image encoders, respectively.
The softmax function between an arbitrary feature $g$ and a set of features $\left\{  f_j \right\}_{j=1}^{C}$ is defined as Eq.~\ref{eq:softmax}.

\begin{equation} \label{eq:softmax}
    \text{softmax} \left( g, \left\{  f_j \right\}_{j=1}^{C} \right)_i  = \frac{\exp(\textrm{sim}(g, f_i))}{\sum_{j} \exp(\textrm{sim}(g, f_j))},
\end{equation}


where $\text{sim}$ indicates cosine similarity between the two features. 
Besides zero-shot classification, CLIP has been utilized in tasks ranging from zero-shot image segmentation~\cite{sam} to a plethora of others~\cite{glip,glipv2}. 

In addition to CLIP, various other VLMs have been proposed~\cite{imagebind,align,albef}.
While maintaining the fundamental concept of training a multi-modal encoder through contrastive learning, efforts have been made to enhance models beyond CLIP.
These include injecting hierarchical embedding structures and incorporating an understanding of negation, as seen in discussions around MeRU~\cite{meru} and CLIPN~\cite{clipn}.



\subsection{Industrial Image Anomaly Detection}
Early works in industrial image anomaly detection that incorporate deep learning often rely on the autoencoder reconstruction loss strategy.
This approach is based on the idea that training an autoencoder (AE) on a dataset consisting mostly of normal data will result in the AE reconstructing normal data well but struggling to reconstruct abnormal data. Some works have added adversarial components to the AE framework~\cite{sabokrou2018adversarially,yan2021learning}, while others have proposed iterative reconstruction methods to maximize the separation between normal and abnormal data~\cite{dehaene2020iterative}. Moreover, in light of the discovery that AEs also reconstruct unseen anomalies well, some studies have introduced negative mining to suppress this capability~\cite{perera2019ocgan}.
RIAD~\cite{zavrtanik2021reconstruction} trains a reconstruction model using an inpainting objective, and there are approaches based on diffusion methods~\cite{wyatt2022anoddpm}.

Another popular approach besides reconstruction loss is feature bank-based methods.
These methods involve creating a feature bank from the training dataset and performing nearest neighbor search on the feature bank to estimate distance-based anomaly scores for query images.
Starting with SPADE~\cite{cohen2020sub} and Patch SVDD~\cite{yi2020patch}, various works have proposed to use features extracted from deep learning models for image anomaly detection.
The choice of feature extractor is a design consideration in feature bank-based approaches, with options including pretrained networks~\cite{cohen2020sub,roth2022towards} or training a feature extractor using self-supervised objective~\cite{yi2020patch} and contrastive objective~\cite{ucad}.

Moreover, variant anomaly detection scenarios such as zero-shot~\cite{jeong2023winclip}, few-shot~\cite{anomalygpt}, and continual anomaly detection~\cite{ucad,inctrl} are also being explored, aiming to push the boundaries of anomaly detection techniques further.

\subsection{VLMs in Industrial Image Anomaly Detection}

Leveraging pre-trained VLMs that have been trained on extensive datasets offers a powerful strategy for industrial image anomaly detection.
The most straightforward application involves using these models as feature extractors~\cite{roth2022towards}.
However, more recent methods exploit the multi-modal capabilities of VLMs to a greater extent.
A prominent example is WinCLIP~\cite{jeong2023winclip}, which harnesses the zero-shot classification ability of CLIP, detailed in Eq.~\ref{eq:clip}, combined with a prompt ensemble technique to enable zero-shot anomaly detection.
The authors enhance the diversity of text prompts by categorizing them into state descriptions and template prompts, effectively delineating both normal and abnormal states. 
Moreover, WinCLIP demonstrates proficiency in few-shot anomaly detection by integrating the results from zero-shot and few-shot based anomaly detection methods.
Although WinCLIP capitalizes on the intermediate layers of the Vision Transformer-based CLIP~\cite{vit}, comparing tokens from these layers in the language space, it is hampered by the misalignment of these tokens with the language space.
APRIL-GAN~\cite{aprilgan} addresses this by introducing a projection layer before comparing patch tokens with text features, thus enhancing performance.

APRIL-GAN generates an anomaly map by comparing the features of text prompts describing normal and abnormal states with the features of a query image, as illustrated in Fig.~\ref{fig:clip_and_coop}(b).
Given sets of text prompts describing normal and abnormal states denoted as $\mathcal{S}{nor}$ and $\mathcal{S}{abn}$, the means of their extracted features are shown in Eq.~\ref{eq:aprilgan_features}.

\begin{equation} \label{eq:aprilgan_features}
\begin{aligned}
    f_{abn} = \frac{1}{| \mathcal{S}_{abn} |} \sum_{s \in  \mathcal{S}_{abn}} E_T (s), \\
    f_{nor} = \frac{1}{| \mathcal{S}_{nor} |} \sum_{s \in  \mathcal{S}_{nor}} E_T (s).
\end{aligned}
\end{equation}

These features are then compared with the intermediate feature map of the image encoder to produce an anomaly map.
More specifically, an anomaly map generated using the $l$-th layer feature map, $\mathbf{A}^l$, can be expressed as shown in Eq.~\ref{eq:aprilgan}.

\begin{equation} \label{eq:aprilgan}
    \mathbf{A}^l_{i,j} = \text{softmax} \left( E^l_I(\mathbf{I})_{i,j}, \left\{ f_{abn}, f_{nor} \right\}   \right)_1,
\end{equation}

where $i$ and $j$ represent the spatial indices of the anomaly map.
Layer-wise calculated anomaly maps are summed to constitute the final anomaly map, \ie, $\textbf{A}_{i,j} = \text{sum}_l \mathbf{A}^l_{i,j}$.

Another recent advancements include proposals that depart from traditional anomaly score-based methods, opting instead for approaches that leverage large language models (LLM)~\cite{anomalygpt}.
Additionally, there has been growing interest in continual anomaly detection methods~\cite{inctrl,ucad}, which aim to train a single anomaly detection model for multiple classes of images.

\section{Methodology}
\subsection{Problem formulation}
First, we provide a formal description of the image anomaly detection problem.
For a given query image, $\mathbf{I}$, the image anomaly detection problem involves quantifying how much $\mathbf{I}$ deviates from normality and computing its anomaly score, $a$.
Typical approaches~\cite{roth2022towards} generate an anomaly map, $\mathbf{A}$, representing the abnormality at each position of the image, and the maximum value of this map is used as the anomaly score, \ie, $\text{max}_{i,j} (\mathbf{A}) = a$.
To achieve this, anomaly detection methods train an anomaly score estimator function, denoted as $\mathcal{A}_\theta (\mathbf{I}) = \mathbf{A}$, where the parameters $\theta$ are trained using a training dataset, $\mathbf{D_{tr}}$.

Each image $\mathbf{I}$ is associated with a label $y$, where a high anomaly score $a$ should be assigned to abnormal data with $y = 1$, and a low score is assigned to normal data with $y = 0$.
The performance of $\mathcal{A}_\theta$ is evaluated using the AUROC (Area Under Receiver Operating Characteristic curve).
The composition of $\mathbf{D_{tr}}$, the training data for the proposed method, can vary depending on the type of supervision employed to train the method.
For example, if training is conducted solely with normal data, the set of normal images $ \mathbf{D_{tr}} = \left\{ \textbf{I}^i_{tr} \right\} $ would be used.
If image-level annotations are available, it would be represented as $ \mathbf{D_{tr}} = \left\{ (\textbf{I}^i_{tr}, y^i) \right\} $.

\begin{figure*}[t]
    \centering
    \includegraphics[width=\linewidth]{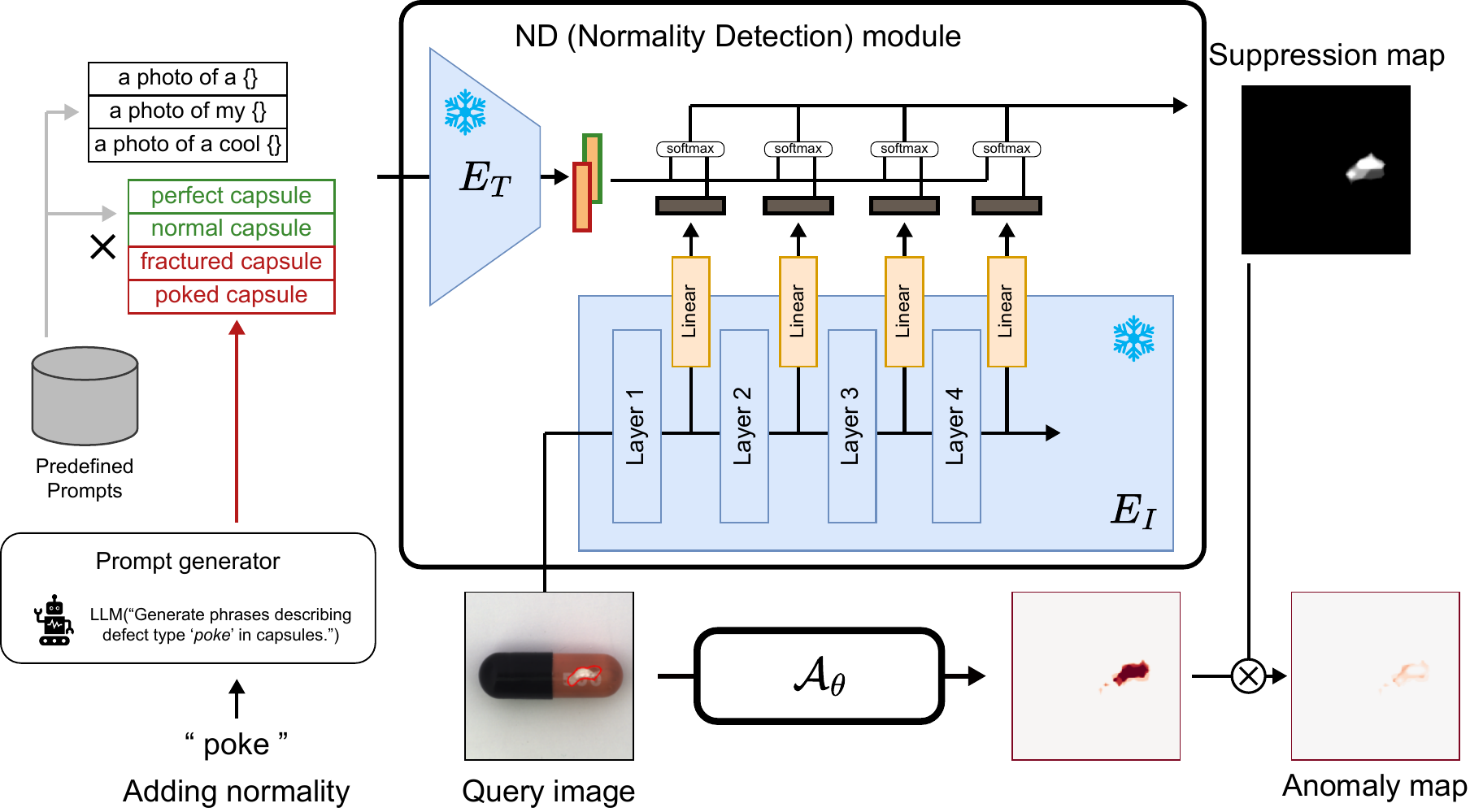}    
    \caption{\textbf{A depiction of the proposed method, NAND.} For a given anomaly detection model ($\mathcal{A}_\theta$) and a text-based normality, NAND starts by generating text prompts for the inputs to the ND module. The output of ND module, a suppression map, is element-wise multiplied to suppress the output of the given anomaly detection model.}
    \label{fig:our_method}
\end{figure*}

In this study, our goal is to incorporate normality addition into existing image anomaly detection models, denoted as $\mathcal{A}_\theta$.
This involves modifying the decision boundary of $\mathcal{A}_\theta$ to adapt to shifts in normality that may occur at test time.
We assume the provision of guidance in the form of text, denoted as $t$, and refer to the modification in the anomaly detection model as \textit{normality addition}.
As a result of normality addition, we develop a new anomaly detection model, $\mathcal{A}_\theta'$, which processes information corresponding to $t$ as normal. This is described in Eq.~\ref{eq:na_problem}.

\begin{equation} \label{eq:na_problem}
    \texttt{addNormality}(\mathcal{A}_\theta, t) = \mathcal{A}_\theta'.
\end{equation}

The resulting anomaly score estimator, $\mathcal{A}_\theta'$, must output anomaly scores that are aligned with the modified definition of normality.
An example of normality addition is presented in Fig.~\ref{fig:teaser}, which demonstrates the anomaly detection model processing the term ``thread'' as normal, using $t=$``thread''.

\subsection{Normality Addition via Normality Detection (NAND)}

Our methodology, termed Normality Addition via Normality Detection (NAND), leverages a process called normality detection to implement normality addition.
This process identifies regions within a query image that correspond to a specified normality $t$, and subsequently suppresses the anomaly scores in these regions in the results of the existing anomaly detection model.
This approach is a post-hoc method, applicable to arbitrary anomaly detection method that generate anomaly maps.

\subsubsection{Prompt Generator}
NAND starts with generating prompts corresponding to the normality for the text encoders of CLIP, using a GPT-based LLM~\cite{gpt-v4}.
We direct the LLM to produce a concise descriptions related to $t$.
For instance, to incorporate the normality $t=$``poke'' for the capsule image class, the prompt to the LLM can be: ``Generate concise phrases describing defects of type `poke' in capsules.''
The resulting phrases, denoted as $\mathcal{T}$, can be $\mathcal{T} = \left\{ \text{``fractured capsule''}, \text{``poked capsule''} \right\}$.
These generated phrases are then integrated as state-level prompts~\cite{jeong2023winclip,aprilgan} of APRIL-GAN, combined with various template-level prompts, to constitute a set of text prompts, $\mathcal{S}_{add}$.
The features of prompts in $\mathcal{S}_{add}$ are extracted using the CLIP text encoder (illustrated in Eq.~\ref{eq:aprilgan_features}) and their mean becomes the representative feature of normality of type \textit{poke}, which is denoted as $f_{add}$.

\begin{figure*}[t]
    \centering
    \includegraphics[width=\linewidth]{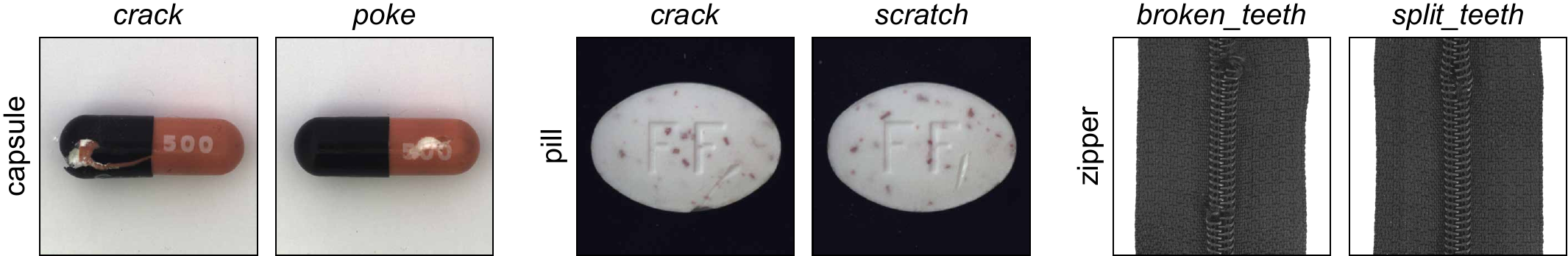}    
    \caption{\textbf{Examples of similar anomaly types in MVTec AD~\cite{mvtecad} dataset.}}    
    \label{fig:similar_anomalytypes}
\end{figure*}
\begin{table*}[t]
\centering
\renewcommand{\arraystretch}{0.8}
\centering
\begin{tabular}{l|cl}
Class      & Anomaly group        & Anomaly types  \\
\hline
\hline
bottle     & broken         & broken\_small \\
& & broken\_large                       \\
           & contamination        & contamination                                    \\
\hline
cable      & missing\_wire         & missing\_wire                                     \\
           & cut\_insulation & cut\_inner\_insulation \\
           & & cut\_outer\_insulation       \\
           & & poke\_insulation     \\
           & missing\_cable        & missing\_cable                                    \\
           & cable\_swap           & cable\_swap                                       \\
           & bent\_wire            & bent\_wire                                        \\
\hline
capsule    & scratch              & scratch                                          \\
           & squeeze              & squeeze                                          \\
           & crack                & crack \\
           & & poke                                      \\
           & faulty\_imprint       & faulty\_imprint                                   \\
\hline
carpet     & thread               & thread                                           \\
           & metal  & metal\_contamination                              \\
           & color                & color                                            \\
           & cut                  & cut \\
           & & hole                                        \\
\hline
grid       & thread               & thread                                           \\
           & bent                 & bent                                             \\
           & glue                 & glue                                             \\
           & metal  & metal\_contamination                              \\
           & broken               & broken                                           \\
\hline
hazelnut   & print                & print                                            \\
           & cut                 & cut \\
           & hole                 & hole \\
           & & crack                                      \\
\hline
leather    & poke                 & poke                                             \\
           & glue                 & glue                                             \\
           & color                & color                                            \\
           & fold                 & fold                                             \\
           & cut                  & cut                                              \\
           
\hline
\end{tabular}
\qquad
\begin{tabular}{l|cl}
Class      & Anomaly group        & Anomaly types                        \\
\hline
\hline
metal\_nut & bent                 & bent                                             \\
           & scratch              & scratch                                          \\
           & color                & color                                            \\
           & flip                 & flip                                             \\
\hline
pill       & pill\_type            & pill\_type                                        \\
           & color                & color                                            \\
           & crack                & crack \\
           & & scratch                                   \\
           & faulty\_imprint       & faulty\_imprint                                   \\
           & contamination        & contamination                                    \\
\hline
screw      & scratch\_head         & scratch\_head                                     \\
           & scratch\_neck         & scratch\_neck                                     \\
           & manipulated\_front    & manipulated\_front                                \\
           & thread           & thread\_top \\
           & & thread\_side                          \\
\hline
tile       & oil                  & oil                                              \\
           & gray\_stroke          & gray\_stroke                                      \\
           & rough                & rough                                            \\
           & crack                & crack                                            \\
           & glue\_strip           & glue\_strip                                       \\
\hline
transistor & misplaced            & misplaced                                        \\
           & damaged\_case         & damaged\_case                                     \\
           & cut\_lead             & cut\_lead                                         \\
           & bent\_lead            & bent\_lead                                        \\
\hline
wood       & scratch              & scratch                                          \\
           & liquid               & liquid                                           \\
           & color                & color                                            \\
           & hole                 & hole                                             \\
\hline
zipper     & fabric        & fabric\_border \\
& & fabric\_interior                   \\
           & teeth         & broken\_teeth \\
           & & squeezed\_teeth \\
           & & split\_teeth \\
           & & rough \\
\hline
\end{tabular}
\vspace{10pt}
\caption{\textbf{Grouped anomaly types in MVTec AD~\cite{mvtecad} dataset.}}\label{table:anomaly_types}
\end{table*}

\subsubsection{Normality Detection Module}
The core of NAND, Normality Detection (ND) module, employs APRIL-GAN~\cite{aprilgan} to detect image regions corresponding to the adding normality.
In the zero-shot anomaly detection process of APRIL-GAN, as shown in Eq.~\ref{eq:aprilgan}, by using $f_{add}$ instead of $f_{abn}$, we can generate a map indicating the presence of patterns related to $t$.
This resulting map, denoted as $\mathbf{A}_{sup}$, represents regions within the query image that correspond to $t$.
Since this map is used to suppress the anomaly map generated by $\mathcal{A}_\theta$, we refer to it as the suppression map.
The suppression map is element-wise multiplied with the output map to produce the final anomaly map, as shown in Eq.~\ref{eq:suppression}.

\begin{equation} \label{eq:suppression}
  \mathbf{A}_{final} = \mathbf{A} \odot (1 - \mathbf{A}_{sup}),
\end{equation}
where $\odot$ denotes element-wise multiplication.
Upon obtaining the final anomaly map, the anomaly score is determined by taking its maximum value, \ie, $a = \text{max}_{i, j} (\mathbf{A}_{final})$.

\section{Experiments}
\subsection{Experimental Setup}
To evaluate the performance of normality addition, we devise a scenario using the MVTec AD benchmark dataset for industrial image anomaly detection~\cite{mvtecad}.
The MVTec AD dataset comprises various anomaly types for each class.
For example, the carpet class includes anomalies such as \textit{color}, \textit{hole}, and \textit{cut}.
We adapt an anomaly detection model trained on the carpet class, $\mathcal{A}_\theta$, to process each anomaly type as normal.
We measure its performance with modified labels by re-defining each anomaly type as normal.
The performance is measured using AUROC, and the adapted anomaly detection model should modify its anomaly score properly to achieve high AUROC.

\begin{figure*}[t]
    \centering
    \includegraphics[width=\linewidth]{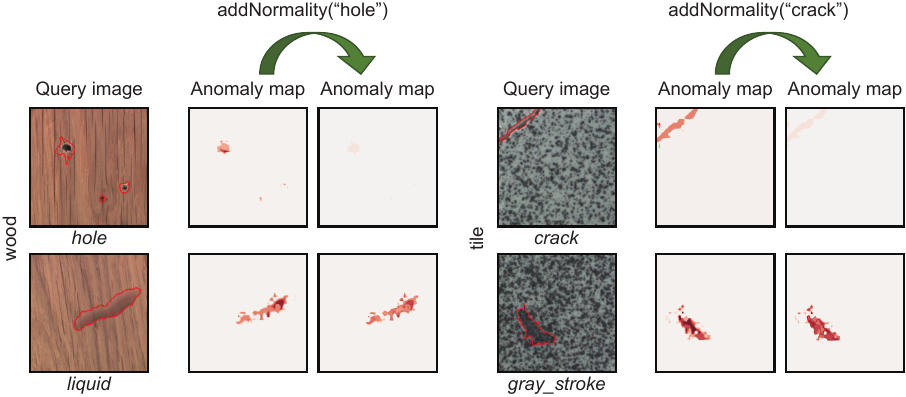}    
    \caption{\textbf{Anomaly maps for input images before and after applying NAND to the anomaly detection model.}}
    \label{fig:examples}
\end{figure*}
\begin{table*}[h]
\centering
\caption{\textbf{Normality addition performance of NAND on MVTec AD~\cite{mvtecad} dataset.}} 
\label{table:auroc}
\renewcommand{\arraystretch}{1.2}
\resizebox{\textwidth}{!}{%
\begin{tabular}{l||l|l|l|l|l}
\hline
Image class     &  \multicolumn{5}{l}{Anomaly type}\\
Average AUROC      & \multicolumn{5}{l}{AUROC}                      \\
\hline
\hline
bottle     & \multicolumn{1}{l|}{broken}      & \multicolumn{1}{l|}{contamination}   &                                        &           &                      \\
56.6 $\rightarrow$ 75.2 (+18.6)       & 23.5    $\rightarrow$ 61.4 (+37.9)          & 89.8 $\rightarrow$ 89.0 (-0.8)                   &                    &           &                      \\
\hline
cable      & \multicolumn{1}{l|}{bent\_wire}         & \multicolumn{1}{l|}{cable\_swap}     & \multicolumn{1}{l|}{cut\_insulation} & \multicolumn{1}{l|}{missing\_wire}        & \multicolumn{1}{l}{missing\_cable} \\
62.6 $\rightarrow$ 64.9  (+2.3) & 53.0 $\rightarrow$ 56.9 (+3.9)       & 68.9 $\rightarrow$ 69.1 (+0.2)          & 57.7 $\rightarrow$ 57.9 (+0.2)         & 66.6 $\rightarrow$ 71.6 (+5.0)        & 66.6 $\rightarrow$ 68.8 (+2.2)          \\
\hline
capsule    & \multicolumn{1}{l|}{crack}          & \multicolumn{1}{l|}{faulty\_imprint} & \multicolumn{1}{l|}{scratch}            & \multicolumn{1}{l|}{squeeze}                  \\
 62.6 $\rightarrow$ 64.7 (+2.1) & 48.0 $\rightarrow$ 56.9 (+8.9)       & 60.5 $\rightarrow$ 64.2 (+3.7)          & 62.4 $\rightarrow$ 60.2 (-2.2)         & 79.4 $\rightarrow$ 77.5 (-1.9)                     \\
\hline
carpet     & \multicolumn{1}{l|}{color}          & \multicolumn{1}{l|}{cut}         & \multicolumn{1}{l|}{metal}   & \multicolumn{1}{l|}{thread}                        \\
77.9 $\rightarrow$ 80.8 (+2.9) & 91.6 $\rightarrow$ 92.2 (+0.6)       & 77.0 $\rightarrow$ 69.5 (-7.5)          & 69.4 $\rightarrow$ 78.2 (+8.8)         & 73.6 $\rightarrow$ 83.5 (+9.9)               \\
\hline
grid       & \multicolumn{1}{l|}{bent}           & \multicolumn{1}{l|}{broken}          & \multicolumn{1}{l|}{glue}               & \multicolumn{1}{l|}{metal} & \multicolumn{1}{l}{thread}           \\
81.1 $\rightarrow$ 82.6 (+1.5) & 88.1 $\rightarrow$ 85.2 (-2.9)       & 69.2 $\rightarrow$ 75.8 (+6.6)          & 83.4 $\rightarrow$ 87.2 (+3.8)         & 74.5 $\rightarrow$ 79.6 (+5.1)        & 90.2 $\rightarrow$ 85.0 (-5.2)          \\
\hline
hazelnut   & \multicolumn{1}{l|}{cut}            & \multicolumn{1}{l|}{hole}        & \multicolumn{1}{l|}{print}              &             &                            \\
80.2 $\rightarrow$ 81.2 (+1.0) & 81.7 $\rightarrow$ 77.0 (-4.7)       & 78.3 $\rightarrow$ 72.1 (-6.2)          & 80.5 $\rightarrow$ 94.4 (+13.9)        &             &                              \\
\hline
leather    & \multicolumn{1}{l|}{color}          & \multicolumn{1}{l|}{cut}         & \multicolumn{1}{l|}{fold}               & \multicolumn{1}{l|}{glue}             & \multicolumn{1}{l}{poke}             \\
81.5 $\rightarrow$ 86.7 (+5.2) & 70.9 $\rightarrow$ 93.7 (+22.8)          & 77.2 $\rightarrow$ 86.6 (+9.4)          & 94.8 $\rightarrow$ 80.1 (-14.7)        & 82.6 $\rightarrow$ 82.5 (-0.1)        & 82.2 $\rightarrow$ 90.6 (+8.4)          \\
\hline
metal\_nut & \multicolumn{1}{l|}{bent}           & \multicolumn{1}{l|}{color}       & \multicolumn{1}{l|}{flip}               & \multicolumn{1}{l|}{scratch}          &                   \\
66.6 $\rightarrow$ 69.4 (+2.8) & 62.6 $\rightarrow$ 67.2 (+4.6)       & 41.9 $\rightarrow$ 52.7 (+10.8)         & 95.8 $\rightarrow$ 95.9 (+0.1)         & 65.9 $\rightarrow$ 61.8 (-4.1)        &                    \\
\hline
pill       & \multicolumn{1}{l|}{color}          & \multicolumn{1}{l|}{contamination}   & \multicolumn{1}{l|}{crack}              & \multicolumn{1}{l|}{faulty\_imprint}      & \multicolumn{1}{l}{pill\_type}       \\
62.4 $\rightarrow$ 69.7 (+7.3) & 52.4 $\rightarrow$ 73.7 (+21.3)          & 63.4 $\rightarrow$ 71.2 (+7.8)          & 63.1 $\rightarrow$ 69.1 (+6.0)         & 56.4 $\rightarrow$ 52.6 (-3.8)        & 76.9 $\rightarrow$ 81.7 (+4.8)          \\
\hline
screw      & \multicolumn{1}{l|}{manipulated\_front} & \multicolumn{1}{l|}{scratch\_head}   & \multicolumn{1}{l|}{scratch\_neck}          & \multicolumn{1}{l|}{thread}          &                 \\
65.1 $\rightarrow$ 67.0 (+1.9) & 66.0 $\rightarrow$ 67.3 (+1.3)       & 73.7 $\rightarrow$ 70.3 (-3.4)          & 62.1 $\rightarrow$ 65.7 (+3.6)         & 58.6 $\rightarrow$ 64.5 (+5.9)        &                   \\
\hline
tile       & \multicolumn{1}{l|}{crack}          & \multicolumn{1}{l|}{glue\_strip}     & \multicolumn{1}{l|}{gray\_stroke}       & \multicolumn{1}{l|}{oil}              & \multicolumn{1}{l}{rough}            \\
80.1 $\rightarrow$ 84.9 (+4.8) & 84.5 $\rightarrow$ 91.5 (+7.0)       & 73.0 $\rightarrow$ 71.9 (-1.1)          & 86.1 $\rightarrow$ 86.7 (+0.6)         & 67.5 $\rightarrow$ 79.8 (+12.3)       & 89.4 $\rightarrow$ 94.4 (+5.0)          \\
\hline
transistor & \multicolumn{1}{l|}{bent\_lead}         & \multicolumn{1}{l|}{cut\_lead}       & \multicolumn{1}{l|}{damaged\_case}          & \multicolumn{1}{l|}{misplaced}        &                 \\
66.6 $\rightarrow$ 70.7 (+4.1) & 64.4 $\rightarrow$ 66.6 (+2.2)       & 60.1 $\rightarrow$ 68.8 (+8.7)          & 65.7 $\rightarrow$ 72.2 (+6.5)         & 76.2 $\rightarrow$ 75.2 (-1.0)        &                           \\
\hline
wood       & \multicolumn{1}{l|}{color}          & \multicolumn{1}{l|}{hole}        & \multicolumn{1}{l|}{liquid}             & \multicolumn{1}{l|}{scratch}          &                    \\
78.7 $\rightarrow$ 84.4 (+5.7)  & 78.7 $\rightarrow$ 87.1 (+8.4)       & 80.8 $\rightarrow$ 91.8 (+11.0)         & 84.6 $\rightarrow$ 83.1 (-1.5)         & 70.8 $\rightarrow$ 75.7 (+4.9)        &                      \\
\hline
zipper     & \multicolumn{1}{l|}{teeth}      & \multicolumn{1}{l|}{fabric}  &              &                     &                            \\
67.0 $\rightarrow$ 68.6 (+1.6)  & 76.0 $\rightarrow$ 80.2 (+4.2)       & 57.9 $\rightarrow$ 56.9 (-1.0)          &              &                     &                    \\
\hline
\end{tabular}}
\end{table*}

However, as shown in Fig.~\ref{fig:similar_anomalytypes}, some anomaly types of MVTec AD dataset within some classes are similar to each other.
We grouped similar anomaly types and readjusted the categorization of anomaly types, as presented in Table~\ref{table:anomaly_types}.
The groups defined in Table~\ref{table:anomaly_types}, referred to as \textit{anomaly groups}, were used as the units for injecting normality addition.
In addition, we excluded toothbrush class from the experiments because it has only one anomaly type, \textit{defective}, and adding this anomaly type as normality makes test dataset all-normal.
Additionally, the cable, pill, wood, and zipper classes have an anomaly type named \textit{combined}, which encompasses all other types of anomalies.
Therefore, the \textit{combined} anomaly type is also excluded from the experiments.

For the backbone of the ND module, APRIL-GAN~\cite{aprilgan}, we utilized the official release by the authors\footnote{\url{https://github.com/ByChelsea/VAND-APRIL-GAN}}.
We generated suppression maps with size of 256 $\times$ 256, and the existing anomaly detection models used in the normality addition experiments were APRIL-GAN~\cite{aprilgan}, WinCLIP~\cite{jeong2023winclip}, and PatchCore~\cite{roth2022towards}.

\subsection{Experimental Results}
We applied the NAND method to various anomaly detection models.
The performance of the AUROC metric before and after implementing NAND, along with the extent of performance improvement, is presented in Table~\ref{table:auroc}, when applied to APRIL-GAN~\cite{aprilgan}.
The results show that AUROC metric increases across various anomaly types, demonstrating that NAND successfully performs normality addition.
On average, NAND improved the AUROC performance from 70.6 to 75.1.

\begin{table}[h]
\centering
\caption{\textbf{Normality addition performance of NAND when applied to various anomaly detection models.}}
\label{table:auroc2}
\renewcommand{\arraystretch}{1.2}
\begin{tabular}{l|cc}
 & \multicolumn{2}{c}{AUROC}  \\
Methods & Before & After \\
\hline
APRIL-GAN~\cite{zhang2019partially} & 70.6     & \textbf{75.1}    \\
WinCLIP-0~\cite{jeong2023winclip} & 65.9     & \textbf{69.9}    \\
WinCLIP-1 & 67.4     & \textbf{71.0}    \\
WinCLIP-5 & 70.4     & \textbf{74.2}    \\
PatchCore~\cite{roth2022towards} & 72.3     & \textbf{75.5}    \\
\end{tabular}
\end{table}

Additionally, the results of applying normality addition to WinCLIP~\cite{jeong2023winclip} and PatchCore~\cite{roth2022towards} are shown in Table~\ref{table:auroc2}.
The WinCLIP-\{0,1,5\} models indicate zero-shot, 1-shot, and 5-shot models, respectively.
The increase in AUROC demonstrates that normality addition is feasible using NAND for commonly used state-of-the-art methods in industrial image anomaly detection.

Fig.~\ref{fig:examples} provides various examples of anomaly maps before and after applying NAND.
The anomaly maps show that the types of anomaly we add are well suppressed, while other types of anomalies are left almost unaffected.
Fig.~\ref{fig:examples}(a) shows that when ``hole'' normality is added to a wood image, the regions corresponding to holes are suppressed in the anomaly score and treated as normal.
However, the image below with liquid stains shows little change.
Additionally, Fig.~\ref{fig:examples}(b) demonstrates that only the anomaly scores corresponding to cracks in the tile image are well suppressed.
Thus, NAND selectively treats only the normality we wish to add as normal, while maintaining other types of anomalies as anomalous.
This indicates successful modification of the decision boundaries through NAND.

\section{Conclusion}
This study introduced a novel approach, Normality Addition via Normality Detection (NAND), and established a scenario of normality addition along with the corresponding evaluation protocol in industrial image anomaly detection.
By leveraging textual descriptions with vision-language models, NAND effectively adjusts traind anomaly detection models to align with new introduction of normality.
Our empirical results on the MVTec AD dataset not only validate the feasibility of normality addition but also demonstrate that NAND successfully achieves normality addition.
The proposed method and scenario hold important ramifications for the practical use of image anomaly detection, particularly in areas where the definition of normality is subject to change due to variations in operational conditions and criteria for quality control.

\begin{acks}
This work was supported by the BK21 FOUR program of the Education and Research Program for Future ICT Pioneers, Seoul National University in 2024, Institute of Information \& Communications Technology Planning \& Evaluation (IITP) grant funded by the Korea government (MSIT) [NO.2021-0-01343, Artificial Intelligence Graduate School Program (Seoul National University)], the National Research Foundation of Korea (NRF) grant funded by the Korea government (MSIT) (No. 2022R1A3B1077720), and Samsung Electronics.
\end{acks}

\bibliographystyle{ACM-Reference-Format}
\bibliography{egbib}


\begin{thebibliography}{27}


\ifx \showCODEN    \undefined \def \showCODEN     #1{\unskip}     \fi
\ifx \showDOI      \undefined \def \showDOI       #1{#1}\fi
\ifx \showISBNx    \undefined \def \showISBNx     #1{\unskip}     \fi
\ifx \showISBNxiii \undefined \def \showISBNxiii  #1{\unskip}     \fi
\ifx \showISSN     \undefined \def \showISSN      #1{\unskip}     \fi
\ifx \showLCCN     \undefined \def \showLCCN      #1{\unskip}     \fi
\ifx \shownote     \undefined \def \shownote      #1{#1}          \fi
\ifx \showarticletitle \undefined \def \showarticletitle #1{#1}   \fi
\ifx \showURL      \undefined \def \showURL       {\relax}        \fi
\providecommand\bibfield[2]{#2}
\providecommand\bibinfo[2]{#2}
\providecommand\natexlab[1]{#1}
\providecommand\showeprint[2][]{arXiv:#2}

\bibitem[Bergmann et~al\mbox{.}(2019)]%
        {mvtecad}
\bibfield{author}{\bibinfo{person}{Paul Bergmann}, \bibinfo{person}{Michael Fauser}, \bibinfo{person}{David Sattlegger}, {and} \bibinfo{person}{Carsten Steger}.} \bibinfo{year}{2019}\natexlab{}.
\newblock \showarticletitle{MVTec AD--A comprehensive real-world dataset for unsupervised anomaly detection}. In \bibinfo{booktitle}{\emph{Proceedings of the IEEE/CVF conference on computer vision and pattern recognition}}. \bibinfo{pages}{9592--9600}.
\newblock


\bibitem[Chen et~al\mbox{.}(2023)]%
        {aprilgan}
\bibfield{author}{\bibinfo{person}{Xuhai Chen}, \bibinfo{person}{Yue Han}, {and} \bibinfo{person}{Jiangning Zhang}.} \bibinfo{year}{2023}\natexlab{}.
\newblock \showarticletitle{A zero-/few-shot anomaly classification and segmentation method for cvpr 2023 vand workshop challenge tracks 1\&2: 1st place on zero-shot ad and 4th place on few-shot ad}.
\newblock \bibinfo{journal}{\emph{arXiv preprint arXiv:2305.17382}} (\bibinfo{year}{2023}).
\newblock


\bibitem[Cohen and Hoshen(2020)]%
        {cohen2020sub}
\bibfield{author}{\bibinfo{person}{Niv Cohen} {and} \bibinfo{person}{Yedid Hoshen}.} \bibinfo{year}{2020}\natexlab{}.
\newblock \showarticletitle{Sub-image anomaly detection with deep pyramid correspondences}.
\newblock \bibinfo{journal}{\emph{arXiv preprint arXiv:2005.02357}} (\bibinfo{year}{2020}).
\newblock


\bibitem[Dehaene et~al\mbox{.}(2020)]%
        {dehaene2020iterative}
\bibfield{author}{\bibinfo{person}{David Dehaene}, \bibinfo{person}{Oriel Frigo}, \bibinfo{person}{S{\'e}bastien Combrexelle}, {and} \bibinfo{person}{Pierre Eline}.} \bibinfo{year}{2020}\natexlab{}.
\newblock \showarticletitle{Iterative energy-based projection on a normal data manifold for anomaly localization}.
\newblock \bibinfo{journal}{\emph{arXiv preprint arXiv:2002.03734}} (\bibinfo{year}{2020}).
\newblock


\bibitem[Desai et~al\mbox{.}(2023)]%
        {meru}
\bibfield{author}{\bibinfo{person}{Karan Desai}, \bibinfo{person}{Maximilian Nickel}, \bibinfo{person}{Tanmay Rajpurohit}, \bibinfo{person}{Justin Johnson}, {and} \bibinfo{person}{Shanmukha~Ramakrishna Vedantam}.} \bibinfo{year}{2023}\natexlab{}.
\newblock \showarticletitle{Hyperbolic image-text representations}. In \bibinfo{booktitle}{\emph{International Conference on Machine Learning}}. PMLR, \bibinfo{pages}{7694--7731}.
\newblock


\bibitem[Dosovitskiy et~al\mbox{.}(2020)]%
        {vit}
\bibfield{author}{\bibinfo{person}{Alexey Dosovitskiy}, \bibinfo{person}{Lucas Beyer}, \bibinfo{person}{Alexander Kolesnikov}, \bibinfo{person}{Dirk Weissenborn}, \bibinfo{person}{Xiaohua Zhai}, \bibinfo{person}{Thomas Unterthiner}, \bibinfo{person}{Mostafa Dehghani}, \bibinfo{person}{Matthias Minderer}, \bibinfo{person}{Georg Heigold}, \bibinfo{person}{Sylvain Gelly}, {et~al\mbox{.}}} \bibinfo{year}{2020}\natexlab{}.
\newblock \showarticletitle{An image is worth 16x16 words: Transformers for image recognition at scale}.
\newblock \bibinfo{journal}{\emph{arXiv preprint arXiv:2010.11929}} (\bibinfo{year}{2020}).
\newblock


\bibitem[Girdhar et~al\mbox{.}(2023)]%
        {imagebind}
\bibfield{author}{\bibinfo{person}{Rohit Girdhar}, \bibinfo{person}{Alaaeldin El-Nouby}, \bibinfo{person}{Zhuang Liu}, \bibinfo{person}{Mannat Singh}, \bibinfo{person}{Kalyan~Vasudev Alwala}, \bibinfo{person}{Armand Joulin}, {and} \bibinfo{person}{Ishan Misra}.} \bibinfo{year}{2023}\natexlab{}.
\newblock \showarticletitle{Imagebind: One embedding space to bind them all}. In \bibinfo{booktitle}{\emph{CVPR}}. \bibinfo{pages}{15180--15190}.
\newblock


\bibitem[Gu et~al\mbox{.}(2024)]%
        {anomalygpt}
\bibfield{author}{\bibinfo{person}{Zhaopeng Gu}, \bibinfo{person}{Bingke Zhu}, \bibinfo{person}{Guibo Zhu}, \bibinfo{person}{Yingying Chen}, \bibinfo{person}{Ming Tang}, {and} \bibinfo{person}{Jinqiao Wang}.} \bibinfo{year}{2024}\natexlab{}.
\newblock \showarticletitle{Anomalygpt: Detecting industrial anomalies using large vision-language models}. In \bibinfo{booktitle}{\emph{Proceedings of the AAAI Conference on Artificial Intelligence}}, Vol.~\bibinfo{volume}{38}. \bibinfo{pages}{1932--1940}.
\newblock


\bibitem[Jeong et~al\mbox{.}(2023)]%
        {jeong2023winclip}
\bibfield{author}{\bibinfo{person}{Jongheon Jeong}, \bibinfo{person}{Yang Zou}, \bibinfo{person}{Taewan Kim}, \bibinfo{person}{Dongqing Zhang}, \bibinfo{person}{Avinash Ravichandran}, {and} \bibinfo{person}{Onkar Dabeer}.} \bibinfo{year}{2023}\natexlab{}.
\newblock \showarticletitle{Winclip: Zero-/few-shot anomaly classification and segmentation}. In \bibinfo{booktitle}{\emph{CVPR}}. \bibinfo{pages}{19606--19616}.
\newblock


\bibitem[Jia et~al\mbox{.}(2021)]%
        {align}
\bibfield{author}{\bibinfo{person}{Chao Jia}, \bibinfo{person}{Yinfei Yang}, \bibinfo{person}{Ye Xia}, \bibinfo{person}{Yi-Ting Chen}, \bibinfo{person}{Zarana Parekh}, \bibinfo{person}{Hieu Pham}, \bibinfo{person}{Quoc Le}, \bibinfo{person}{Yun-Hsuan Sung}, \bibinfo{person}{Zhen Li}, {and} \bibinfo{person}{Tom Duerig}.} \bibinfo{year}{2021}\natexlab{}.
\newblock \showarticletitle{Scaling up visual and vision-language representation learning with noisy text supervision}. In \bibinfo{booktitle}{\emph{ICML}}. PMLR, \bibinfo{pages}{4904--4916}.
\newblock


\bibitem[Kirillov et~al\mbox{.}(2023)]%
        {sam}
\bibfield{author}{\bibinfo{person}{Alexander Kirillov}, \bibinfo{person}{Eric Mintun}, \bibinfo{person}{Nikhila Ravi}, \bibinfo{person}{Hanzi Mao}, \bibinfo{person}{Chloe Rolland}, \bibinfo{person}{Laura Gustafson}, \bibinfo{person}{Tete Xiao}, \bibinfo{person}{Spencer Whitehead}, \bibinfo{person}{Alexander~C Berg}, \bibinfo{person}{Wan-Yen Lo}, {et~al\mbox{.}}} \bibinfo{year}{2023}\natexlab{}.
\newblock \showarticletitle{Segment anything}. In \bibinfo{booktitle}{\emph{Proceedings of the IEEE/CVF International Conference on Computer Vision}}. \bibinfo{pages}{4015--4026}.
\newblock


\bibitem[Li et~al\mbox{.}(2021)]%
        {albef}
\bibfield{author}{\bibinfo{person}{Junnan Li}, \bibinfo{person}{Ramprasaath Selvaraju}, \bibinfo{person}{Akhilesh Gotmare}, \bibinfo{person}{Shafiq Joty}, \bibinfo{person}{Caiming Xiong}, {and} \bibinfo{person}{Steven Chu~Hong Hoi}.} \bibinfo{year}{2021}\natexlab{}.
\newblock \showarticletitle{Align before fuse: Vision and language representation learning with momentum distillation}.
\newblock \bibinfo{journal}{\emph{NeurIPS}}  \bibinfo{volume}{34} (\bibinfo{year}{2021}), \bibinfo{pages}{9694--9705}.
\newblock


\bibitem[Li et~al\mbox{.}(2022)]%
        {glip}
\bibfield{author}{\bibinfo{person}{Liunian~Harold Li}, \bibinfo{person}{Pengchuan Zhang}, \bibinfo{person}{Haotian Zhang}, \bibinfo{person}{Jianwei Yang}, \bibinfo{person}{Chunyuan Li}, \bibinfo{person}{Yiwu Zhong}, \bibinfo{person}{Lijuan Wang}, \bibinfo{person}{Lu Yuan}, \bibinfo{person}{Lei Zhang}, \bibinfo{person}{Jenq-Neng Hwang}, {et~al\mbox{.}}} \bibinfo{year}{2022}\natexlab{}.
\newblock \showarticletitle{Grounded language-image pre-training}. In \bibinfo{booktitle}{\emph{CVPR}}. \bibinfo{pages}{10965--10975}.
\newblock


\bibitem[Liu et~al\mbox{.}(2024)]%
        {ucad}
\bibfield{author}{\bibinfo{person}{Jiaqi Liu}, \bibinfo{person}{Kai Wu}, \bibinfo{person}{Qiang Nie}, \bibinfo{person}{Ying Chen}, \bibinfo{person}{Bin-Bin Gao}, \bibinfo{person}{Yong Liu}, \bibinfo{person}{Jinbao Wang}, \bibinfo{person}{Chengjie Wang}, {and} \bibinfo{person}{Feng Zheng}.} \bibinfo{year}{2024}\natexlab{}.
\newblock \showarticletitle{Unsupervised Continual Anomaly Detection with Contrastively-learned Prompt}.
\newblock \bibinfo{journal}{\emph{arXiv preprint arXiv:2401.01010}} (\bibinfo{year}{2024}).
\newblock


\bibitem[Perera et~al\mbox{.}(2019)]%
        {perera2019ocgan}
\bibfield{author}{\bibinfo{person}{Pramuditha Perera}, \bibinfo{person}{Ramesh Nallapati}, {and} \bibinfo{person}{Bing Xiang}.} \bibinfo{year}{2019}\natexlab{}.
\newblock \showarticletitle{Ocgan: One-class novelty detection using gans with constrained latent representations}. In \bibinfo{booktitle}{\emph{Proceedings of the IEEE/CVF conference on computer vision and pattern recognition}}. \bibinfo{pages}{2898--2906}.
\newblock


\bibitem[Radford et~al\mbox{.}(2021)]%
        {clip2021icml}
\bibfield{author}{\bibinfo{person}{Alec Radford}, \bibinfo{person}{Jong~Wook Kim}, \bibinfo{person}{Chris Hallacy}, \bibinfo{person}{Aditya Ramesh}, \bibinfo{person}{Gabriel Goh}, \bibinfo{person}{Sandhini Agarwal}, \bibinfo{person}{Girish Sastry}, \bibinfo{person}{Amanda Askell}, \bibinfo{person}{Pamela Mishkin}, \bibinfo{person}{Jack Clark}, {et~al\mbox{.}}} \bibinfo{year}{2021}\natexlab{}.
\newblock \showarticletitle{Learning transferable visual models from natural language supervision}. In \bibinfo{booktitle}{\emph{ICML}}. PMLR, \bibinfo{pages}{8748--8763}.
\newblock


\bibitem[Roth et~al\mbox{.}(2022)]%
        {roth2022towards}
\bibfield{author}{\bibinfo{person}{Karsten Roth}, \bibinfo{person}{Latha Pemula}, \bibinfo{person}{Joaquin Zepeda}, \bibinfo{person}{Bernhard Sch{\"o}lkopf}, \bibinfo{person}{Thomas Brox}, {and} \bibinfo{person}{Peter Gehler}.} \bibinfo{year}{2022}\natexlab{}.
\newblock \showarticletitle{Towards total recall in industrial anomaly detection}. In \bibinfo{booktitle}{\emph{Proceedings of the IEEE/CVF Conference on Computer Vision and Pattern Recognition}}. \bibinfo{pages}{14318--14328}.
\newblock


\bibitem[Sabokrou et~al\mbox{.}(2018)]%
        {sabokrou2018adversarially}
\bibfield{author}{\bibinfo{person}{Mohammad Sabokrou}, \bibinfo{person}{Mohammad Khalooei}, \bibinfo{person}{Mahmood Fathy}, {and} \bibinfo{person}{Ehsan Adeli}.} \bibinfo{year}{2018}\natexlab{}.
\newblock \showarticletitle{Adversarially learned one-class classifier for novelty detection}. In \bibinfo{booktitle}{\emph{Proceedings of the IEEE conference on computer vision and pattern recognition}}. \bibinfo{pages}{3379--3388}.
\newblock


\bibitem[Wang et~al\mbox{.}(2023)]%
        {clipn}
\bibfield{author}{\bibinfo{person}{Hualiang Wang}, \bibinfo{person}{Yi Li}, \bibinfo{person}{Huifeng Yao}, {and} \bibinfo{person}{Xiaomeng Li}.} \bibinfo{year}{2023}\natexlab{}.
\newblock \showarticletitle{Clipn for zero-shot ood detection: Teaching clip to say no}. In \bibinfo{booktitle}{\emph{Proceedings of the IEEE/CVF International Conference on Computer Vision}}. \bibinfo{pages}{1802--1812}.
\newblock


\bibitem[Wyatt et~al\mbox{.}(2022)]%
        {wyatt2022anoddpm}
\bibfield{author}{\bibinfo{person}{Julian Wyatt}, \bibinfo{person}{Adam Leach}, \bibinfo{person}{Sebastian~M Schmon}, {and} \bibinfo{person}{Chris~G Willcocks}.} \bibinfo{year}{2022}\natexlab{}.
\newblock \showarticletitle{Anoddpm: Anomaly detection with denoising diffusion probabilistic models using simplex noise}. In \bibinfo{booktitle}{\emph{Proceedings of the IEEE/CVF Conference on Computer Vision and Pattern Recognition}}. \bibinfo{pages}{650--656}.
\newblock


\bibitem[Yan et~al\mbox{.}(2021)]%
        {yan2021learning}
\bibfield{author}{\bibinfo{person}{Xudong Yan}, \bibinfo{person}{Huaidong Zhang}, \bibinfo{person}{Xuemiao Xu}, \bibinfo{person}{Xiaowei Hu}, {and} \bibinfo{person}{Pheng-Ann Heng}.} \bibinfo{year}{2021}\natexlab{}.
\newblock \showarticletitle{Learning semantic context from normal samples for unsupervised anomaly detection}. In \bibinfo{booktitle}{\emph{Proceedings of the AAAI conference on artificial intelligence}}, Vol.~\bibinfo{volume}{35}. \bibinfo{pages}{3110--3118}.
\newblock


\bibitem[Yang et~al\mbox{.}(2023)]%
        {gpt-v4}
\bibfield{author}{\bibinfo{person}{Zhengyuan Yang}, \bibinfo{person}{Linjie Li}, \bibinfo{person}{Kevin Lin}, \bibinfo{person}{Jianfeng Wang}, \bibinfo{person}{Chung-Ching Lin}, \bibinfo{person}{Zicheng Liu}, {and} \bibinfo{person}{Lijuan Wang}.} \bibinfo{year}{2023}\natexlab{}.
\newblock \showarticletitle{The dawn of lmms: Preliminary explorations with gpt-4v (ision)}.
\newblock \bibinfo{journal}{\emph{arXiv preprint arXiv:2309.17421}} \bibinfo{volume}{9}, \bibinfo{number}{1} (\bibinfo{year}{2023}).
\newblock


\bibitem[Yi and Yoon(2020)]%
        {yi2020patch}
\bibfield{author}{\bibinfo{person}{Jihun Yi} {and} \bibinfo{person}{Sungroh Yoon}.} \bibinfo{year}{2020}\natexlab{}.
\newblock \showarticletitle{Patch svdd: Patch-level svdd for anomaly detection and segmentation}. In \bibinfo{booktitle}{\emph{Proceedings of the Asian conference on computer vision}}.
\newblock


\bibitem[Zavrtanik et~al\mbox{.}(2021)]%
        {zavrtanik2021reconstruction}
\bibfield{author}{\bibinfo{person}{Vitjan Zavrtanik}, \bibinfo{person}{Matej Kristan}, {and} \bibinfo{person}{Danijel Sko{\v{c}}aj}.} \bibinfo{year}{2021}\natexlab{}.
\newblock \showarticletitle{Reconstruction by inpainting for visual anomaly detection}.
\newblock \bibinfo{journal}{\emph{Pattern Recognition}}  \bibinfo{volume}{112} (\bibinfo{year}{2021}), \bibinfo{pages}{107706}.
\newblock


\bibitem[Zhang and Hoi(2019)]%
        {zhang2019partially}
\bibfield{author}{\bibinfo{person}{Chen Zhang} {and} \bibinfo{person}{Steven~CH Hoi}.} \bibinfo{year}{2019}\natexlab{}.
\newblock \showarticletitle{Partially Observable Multi-Sensor Sequential Change Detection: A Combinatorial Multi-Armed Bandit Approach}. \bibinfo{pages}{5733--5740}.
\newblock


\bibitem[Zhang et~al\mbox{.}(2022)]%
        {glipv2}
\bibfield{author}{\bibinfo{person}{Haotian Zhang}, \bibinfo{person}{Pengchuan Zhang}, \bibinfo{person}{Xiaowei Hu}, \bibinfo{person}{Yen-Chun Chen}, \bibinfo{person}{Liunian Li}, \bibinfo{person}{Xiyang Dai}, \bibinfo{person}{Lijuan Wang}, \bibinfo{person}{Lu Yuan}, \bibinfo{person}{Jenq-Neng Hwang}, {and} \bibinfo{person}{Jianfeng Gao}.} \bibinfo{year}{2022}\natexlab{}.
\newblock \showarticletitle{Glipv2: Unifying localization and vision-language understanding}.
\newblock \bibinfo{journal}{\emph{NeurIPS}}  \bibinfo{volume}{35} (\bibinfo{year}{2022}), \bibinfo{pages}{36067--36080}.
\newblock


\bibitem[Zhu and Pang(2024)]%
        {inctrl}
\bibfield{author}{\bibinfo{person}{Jiawen Zhu} {and} \bibinfo{person}{Guansong Pang}.} \bibinfo{year}{2024}\natexlab{}.
\newblock \showarticletitle{Toward Generalist Anomaly Detection via In-context Residual Learning with Few-shot Sample Prompts}.
\newblock \bibinfo{journal}{\emph{arXiv preprint arXiv:2403.06495}} (\bibinfo{year}{2024}).
\newblock


\end{thebibliography}

\end{document}